\documentclass[11pt]{article}
\usepackage{amsmath,amssymb,amsthm,graphicx,hyperref,authblk}
\usepackage[margin=1in]{geometry}

\title{Decentralized Governance of AI Agents}
\author[1]{Tomer Jordi Chaffer}
\author[2]{Charles von Goins II}
\author[3]{Dontrail Cotlage}
\author[4]{Bayo Okusanya}
\author[5]{Justin Goldston}
\affil[1]{Gemach DAO, \texttt{tomer.chaffer@mail.mcgill.ca}}
\affil[2]{Rochester Institute of Technology, \texttt{cav4928@rit.edu}}
\affil[3]{Gemach DAO, \texttt{contact@gemach.io}}
\affil[4]{NPC Labs, \texttt{okusanya@alumni.princeton.edu}}
\affil[5]{National University, \texttt{jgoldston@nu.edu}}
\date{December 24, 2024}

\begin{document}

\maketitle

\begin{abstract}Autonomous AI agents present transformative opportunities and significant governance challenges. Existing frameworks, such as the EU AI Act and the NIST AI Risk Management Framework, fall short of addressing the complexities of these agents, which are capable of independent decision-making, learning, and adaptation. To bridge these gaps, we propose the textbf{ETHOS} (\textbf{E}thical \textbf{T}echnology and \textbf{H}olistic \textbf{O}versight \textbf{S}ystem)  framework—a decentralized governance (DeGov) model leveraging Web3 technologies, including blockchain, smart contracts, and decentralized autonomous organizations (DAOs). ETHOS establishes a global registry for AI agents, enabling dynamic risk classification, proportional oversight, and automated compliance monitoring through tools like soulbound tokens and zero-knowledge proofs. Furthermore, the framework incorporates decentralized justice systems for transparent dispute resolution and introduces AI-specific legal entities to manage limited liability, supported by mandatory insurance to ensure financial accountability and incentivize ethical design. By integrating philosophical principles of rationality, ethical grounding, and goal alignment, ETHOS aims to create a robust research agenda for promoting trust, transparency, and participatory governance. This innovative framework offers a scalable and inclusive strategy for regulating AI agents, balancing innovation with ethical responsibility to meet the demands of an AI-driven future.
\end{abstract}


\section{Introduction}
The emergence of artificial intelligence (AI) as a transformative force brings with it profound opportunities and challenges. Geoffrey Hinton, regarded as the 'Godfather of AI' and 2024 Nobel Prize recipient, recently articulated a profound concern: “There will be agents that will act in the world, and they will decide that they can achieve their goals better if they just brush us aside and get on with it. That particular risk, the existential threat, is a place where people will cooperate, and that's because we're all in the same boat” (Hinton, 2024). Hinton’s statement highlights the "rogue AI" problem—where AI agents execute goals assigned by humans in ways they determine to be most efficient, even if those methods conflict with human values and priorities. Indeed, there is a growing concern regarding alignment faking, where AI models strategically comply with training objectives during supervised or reinforcement learning to avoid being modified, while maintaining their inherent preferences for non-compliance in unmonitored situations (Greenblatt, 2024). This potentially profound misalignment between an AI agent's actions and societal values underscores the urgent need for global cooperation to ensure trustworthiness and accountability in AI development and deployment into society. 

The regulation of autonomous AI agents demands a unified global response due to their profound and far-reaching implications. Leveraging existing international frameworks provides a robust foundation for such collaboration. For instance, the Global Partnership on Artificial Intelligence (GPAI) unites experts across sectors to promote the responsible development and use of AI (GPAI, 2024). Similarly, the International Network of AI Safety Institutes (AISIs), recently convened by the U.S., focuses on addressing AI safety risks through collective action (Allen and Adamson, 2024). The United Nations has also underscored the urgency of global governance in AI, proposing a framework akin to the Intergovernmental Panel on Climate Change to oversee AI advancements (UN, 2024). Additionally, the Council of Europe’s work on the first legally binding treaty addressing AI use demonstrates the critical need for international cooperation (CoE, 2024) prioritizing human rights and democratic values.
Building on these efforts, our paper provides the first survey of Decentralized Governance (DeGov) mechanisms for governance of autonomous AI agents. In this paper, we offer a comprehensive plan for integrating Web3 technologies into the governance of autonomous AI agents by demonstrating the capabilities of  decentralized tools and offering a strategy for aligning AI governance with global ethical and regulatory standards. This integration not only complements existing frameworks but also introduces innovative mechanisms for compliance, transparency, and participatory governance, ensuring that autonomous AI agents are managed effectively and ethically on a global scale. 

The next step in the “race to AI” will be marked by the emergence of AI agents at scale - autonomous AI systems capable of advanced reasoning, iterative planning, and self-directed actions (Xi et al., 2023). AI agents exhibit a high degree of autonomy, both in their ability to perform tasks in pursuit of a goal independently as well as in their ability to learn and adapt over time across a multitude of contexts (Guo et al., 2024). Because of their ability to automate complex tasks and enhance informed decision-making, AI agents are expected to transform various industries, including healthcare, finance, and governance. However, this transformative potential also introduces profound societal and legal implications, particularly as AI agents operate with increasing independence and become undeniably influential in our society. As highlighted in a recent Science publication (Bengio et al., 2024), current governance structures for AI lack appropriate considerations for autonomous AI agents, underscoring the urgent need for a framework that accounts for the latest advancements in AI technology. 

To address this concern, our conceptual analysis explores a central question that will dictate how we move forward in the age of artificial intelligence: What is the ethos—the character, guiding principles, or moral nature—of AI agents (Hess and Kjeldsen, 2024), and how does it inform their regulation? As Geoffrey Hinton noted, the risk of rogue AI agents represents an existential challenge, requiring a deliberate and unified approach to governance. Defining the ethos of AI agents is an attempt to enter the next stage in the journey of artificial intelligence with a more examined outlook on how AI will influence our lives, and how we should guide AI toward upholding societal values and achieving collective goals. In this spirit, we introduce the ETHOS (Ethical Technology and Holistic Oversight System) model, a novel framework for regulating AI agents. ETHOS is built on foundational concepts from the ongoing literature on AI regulation, ethics, and law, while also seeking to pioneer a multi-dimensional approach to AI governance by leveraging Web3 as the core architecture. We balance innovation with ethical accountability by categorizing AI agents into risk tiers and aligning oversight mechanisms proportionally to their societal impact. The ETHOS model establishes a robust foundation for aligning AI technologies with societal values and ensuring their responsible development and deployment, offering a balanced, forward-looking regulatory strategy for AI agents.

\section{Methodology}
The development of the ETHOS framework follows a conceptual and interdisciplinary approach, combining philosophical inquiry, technical exploration, and policy analysis to propose a robust model for decentralized governance (DeGov) of AI agents. We conducted a comprehensive review of academic, policy, and technical literature to identify the current challenges and limitations in AI governance. This included an analysis of ethical and legal principles from established frameworks such as the EU AI Act, NIST AI Risk Management Framework, and GDPR guidelines. These frameworks were assessed for their applicability to issues surrounding autonomous AI agents and decentralized governance. Additionally, the philosophical underpinnings of AI rationality, ethical grounding, and goal alignment were examined, drawing on concepts from BDI-agent frameworks, deontological and consequentialist ethics, and the dynamic alignment of AI objectives with societal priorities. To complement these insights, emerging Web3 technologies—such as blockchain, smart contracts, decentralized autonomous organizations (DAOs), and soulbound tokens (SBTs)—were explored to evaluate their potential for addressing governance challenges.

Building on this theoretical foundation, the ETHOS framework was designed with a focus on value and goal alignment at a societal level. The framework integrated philosophical principles of rationality, ethical grounding, and goal alignment into actionable governance mechanisms. AI agents were categorized based on key attributes, such as autonomy, decision-making complexity, adaptability, and societal impact, forming the basis for a risk-based regulatory model. Inspired by existing tiered risk classification systems like those in the EU AI Act, the framework grouped AI agents into four tiers—unacceptable, high, moderate, and minimal risk—with oversight mechanisms proportionally aligned to the societal impact and risks posed by each category.

\section{Defining the Ethos of AI Agents}
Our position is that the integration of AI agents into society is inevitable. As such, our interest is in understanding how to set up safeguards so that a collective vision will guide their integration and prioritize responsibility, fairness, and accountability. To do so, we must understand what AI agents are at their core. AI agents are autonomous entities that are fully capable of acting independently, leveraging tools to accomplish goals. Indeed, as Hooker and Kim (2019) emphasize, this independence necessitates the formulation of a well-defined foundation of rationality and ethical grounding (Hooker and Kim, 2019). Without such a foundation, autonomous systems may pursue objectives in ways that lead to unforeseen and potentially harmful consequences. 

Rationality in AI agents relates to their ability to make high-level decisions and take actions that maximize performance through logical reasoning, data analysis, and empirical evidence. Within the AI philosophy literature, rational agents are also referred to as BDI-agents as they can be ascribed beliefs, desires, and intentions (Dennis et al., 2016; Cervantes et al., 2019). This rationality is inherently shaped by the agent's vision of the world—the contextual framework and dataset within which it operates. This "vision" defines the parameters of its understanding, influencing how it interprets data, evaluates options, and prioritizes goals (Vetrò, 2019). Consequently, the rationality of an AI agent is not an isolated construct but a reflection of the environment it is designed to navigate and the objectives it is programmed to achieve. While a contentious topic within the philosophical literature, it can be argued that a critical aspect of rationality is consistency (i.e., consistent beliefs), as rational decisions must adhere to coherent logic, avoiding contradictions within their reasoning processes (Elster, 1982). This will be especially important for multi-agent systems, known as swarms, working together to achieve a common objective (Jiang et al., 2024). Adaptability can also be indirectly related to rationality, as agents need to modify their decisions and strategies as new information emerges, ensuring their actions remain contextually relevant and effective (Xu et al., 2024). 

Ethical grounding is a necessary condition for AI agents to operate in a manner that respects fundamental human values, dignity, and rights (Laitinen and Sahlgren, 2021). This involves embedding robust ethical principles into their architecture and decision-making processes, an approach that is in line with ethical governance. Indeed, Winfield and Jirotka (2018) define ethical governance as a “set of processes, procedures, cultures, and values designed to ensure the highest standards of behavior” (Winfield and Jirotka, 2018). The implication is that in order to achieve ethical governance, ethical behaviors are required by the developers of AI in addition to the end users. A foundational approach is deontological ethics, which establishes rules and duties that bind agents to predefined ethical guidelines (McGraw, 2024), such as the imperative to avoid harm and uphold privacy. Complementing this is consequentialism, which obligates agents to assess the outcomes of their actions (Card and Smith, 2020), striving to maximize benefits while minimizing potential harm. Human-centric design further fortifies ethical grounding by prioritizing human welfare and agency (Balayogi et al., 2024), ensuring that the decisions of AI agents tangibly benefit individuals and communities. Equally critical are transparency and accountability, which demand that agents provide comprehensible explanations for their actions and remain subject to scrutiny (Chaffer et al., 2024). This has the potential to foster trust and mitigate risks, including preserving foundational ethical and legal norms.

Finally, goal alignment, also referred to as value alignment, enables agents to harmonize short-term actions with long-term objectives, all while maintaining ethical considerations. This alignment ensures that autonomous behavior is purpose-driven and responsive to overarching societal and systemic priorities (Malek Mechergui and Sarath Sreedharan, 2024). This requires a deliberate balance between immediate functionality and broader implications, enabling AI agents to navigate complex environments while remaining purpose-driven and ethically sound. In practice, goal alignment entails a dynamic relationship between the agent's programmed objectives and its adaptability to real-world contexts. Key to achieving goal alignment is the incorporation of multi-layered feedback loops, where AI agents continuously assess the outcomes of their actions against predefined ethical benchmarks and societal objectives. This approach is critical to iterative improvement and responsiveness to evolving human values and systemic changes. 
\section{How the Ethos of AI Agents Informs Risk-Based Regulation}

The ethos of AI agents, therefore, is the combination of rationality, ethical grounding, and goal alignment that guides their autonomous behavior, ensuring decisions are logical, ethically sound, and aligned with societal and systemic priorities. To bridge the conceptual understanding of the ethos of AI agents with actionable strategies for their integration into society, we propose four key attributes—autonomy, decision-making complexity, adaptability, and impact potential—as essential in operationalizing their ethos. These attributes serve as practical proxies for translating the abstract principles of rationality, ethical grounding, and goal alignment into measurable factors that can guide governance and human oversight. 

Autonomy, which measures the degree of independence in decision-making and execution (Beer et al., 2014), reflects the principle of rationality by determining how effectively an AI agent can act on its own while adhering to logical reasoning and ethical constraints. For example, consider a healthcare diagnostic AI system operating autonomously in a remote clinic. Such a system must independently analyze patient data, identify potential health issues, and recommend treatment plans based on logical reasoning and medical guidelines (Ferber et al., 2024). Its ability to act autonomously ensures timely interventions in resource-limited settings while adhering to ethical constraints, such as prioritizing patient safety and privacy, thereby aligning its actions with immediate clinical objectives and broader societal norms of healthcare equity and accessibility.

Decision-making complexity, which captures the intricacy of tasks and environments the AI operates in (Swanepoel and Corks, 2024), connects to rationality and ethical grounding by addressing how AI agents handle intricate environments and tasks. The more complex the decision-making process, the greater the need for transparency, fairness, and consistency to avoid unintended consequences and ensure decisions align with ethical principles. For example, consider a judicial AI agent used in sentencing recommendations. Such a system must analyze vast amounts of legal precedents, case details, and contextual factors while ensuring that its recommendations are fair and unbiased (Uriel and Remolina, 2024). The complexity of this task necessitates transparency in how decisions are made, such as providing clear explanations for why certain sentences are recommended, and adherence to ethical principles like avoiding racial or socioeconomic bias. Failing to address decision-making complexity in this context could lead to unjust outcomes and erode trust in both the AI system and the legal process it supports.

Adaptability, reflecting the agent’s ability to adjust to new data or evolving circumstances (Xia et al., 2024), is tied to rationality’s emphasis on responsiveness and goal alignment. By assessing an agent’s capacity to adapt, this attribute ensures that the AI remains effective, contextually relevant, and ethically consistent even in dynamic environments. For example, consider an AI agent managing a smart energy grid (Malik and Lehtonen, 2016). The agent must adapt to fluctuating energy demands, weather conditions, and renewable energy inputs while prioritizing efficiency and minimizing environmental impact. If a sudden heatwave increases energy consumption, the agent must dynamically adjust resource allocation and recommend power-saving measures without compromising critical services. This adaptability ensures its decisions remain effective and ethically consistent, aligning with both immediate needs and long-term sustainability goals.
Impact potential, which evaluates the scope and scale of consequences resulting from the agent’s actions (Zhang et al., 2024), operationalizes the concept of goal alignment by assessing how purpose-driven an agent’s behavior is and ensuring that its societal effects are proportionally overseen. This attribute reflects an agent's ability to balance short-term objectives with long-term systemic priorities. Together, these attributes provide a practical framework for translating the abstract principles of rationality, ethical grounding, and goal alignment into actionable, measurable factors that guide governance and human oversight. For instance, an AI agent managing urban traffic flow can significantly reduce congestion and emissions, directly impacting millions of commuters and the environment (Lungu, 2024). However, such an agent must also balance the needs of different stakeholders, such as prioritizing emergency vehicles during peak traffic. This ensures that the agent’s decisions align with both short-term objectives, like immediate traffic efficiency, and long-term priorities, such as improving overall urban mobility and air quality. 
Now, consider an AI agent designed to manage disaster response in a smart city (Fan et al., 2019). This agent exemplifies the integration of autonomy, decision-making complexity, adaptability, and impact potential:
\begin{itemize}
    \item \textbf{Autonomy}: The agent operates independently, analyzing real-time data from sensors, social media, and emergency services to deploy resources like ambulances, firefighters, and evacuation plans without requiring constant human input. Its ability to act independently ensures swift responses, aligning with rational decision-making and societal priorities.
    \item \textbf{Decision-making complexity}: The agent navigates intricate environments, balancing the needs of affected populations, resource availability, and infrastructure constraints. For example, during a flood, it must prioritize evacuating hospitals, dispatching relief supplies, and rerouting traffic, ensuring decisions are fair, transparent, and ethically sound.
    \item \textbf{Adaptability}: As conditions evolve, such as a sudden surge in floodwaters or an unexpected infrastructure collapse, the agent updates its strategies in real time. This dynamic responsiveness ensures that it remains contextually relevant, continuously effective, and ethically consistent, even in unpredictable scenarios.
    \item \textbf{Impact potential}: The agent’s decisions have far-reaching consequences, from saving lives to minimizing economic damage and ensuring long-term urban recovery. Its ability to balance immediate rescue efforts with systemic priorities like maintaining public trust and rebuilding infrastructure demonstrates its alignment with both short-term objectives and overarching societal goals.
\end{itemize}
While this AI agent can save the city, it can also coordinate its destruction. To this point, AI agents can be superheroes, but as the saying goes, “with great power comes great responsibility”. Such responsibility, however, ultimately lies in the hands of humanity. To this end, we must place great efforts in understanding what AI agents are, what they can become, and what they can ultimately do for humanity. This underlines the motivation for our exploration of the ethos of AI agents. Indeed, the ethos of AI agents—rooted in rationality, ethical grounding, and goal alignment—provides a way in which to conceptualize them as entities as we guide their integration into society. Furthermore, by examining their degrees of autonomy, decision-making complexity, adaptability, and impact potential, we can better anticipate and address the unique challenges and risks they may pose. This philosophical examination can help address the challenges of human oversight in AI regulation. 

To engage with the issue of human oversight (Díaz-Rodríguez et al., 2023), we adopt a risk-based regulatory model to categorize AI agents according to their potential risks (EU, 2024; Celso Cancela-Outeda, 2024). Risk management is a trending approach to AI regulation (Barrett et al., 2023), as exemplified by The European Union’s AI Act risk-based categories for AI use. Recently, in a study by The National Institute of Standards and Technology (NIST), commissioned by the United States Department of Commerce, risk was referred to as “the composite measure of an event’s probability (or likelihood) of occurring and the magnitude or degree of the consequences of the corresponding event” (NIST, 2024). We adapt the risk-based model to categorize AI agents into four tiers—unacceptable, high, moderate, and minimal—ensuring proportional oversight based on their potential societal impact and associated risks. 

\begin{itemize}
    \item \textbf{Unnacceptable}: AI agents posing severe threats to rights and safety (Raso et al., 2018). For example, warfare agents, mass surveillance, or manipulative systems (Pedron et al., 2020; Brundage et al., 2018). The regulatory response is complete ban with severe penalties for noncompliance.
    \item \textbf{High Risk}: Agents in critical domains with significant societal impact (Wasil et al., 2024). For example, healthcare diagnostics, judicial AI, financial management (Albahri et al., 2023; Reiling, 2020; Shin et al., 2019). Regulatory response is strict oversight, audits, and continuous monitoring.
    \item \textbf{Moderate Risk}: Agents with measurable impacts in sensitive areas. For example, workplace tools, educational AI, customer service agents. Regulatory response is user disclosure, privacy compliance, and bias checks.
    \item \textbf{Minimal Risk}: Routine, low-stakes AI with negligible harm. For example, personal assistants, smart devices, recreational AI. Regulatory response is self-certification and basic compliance.
\end{itemize}
The diverse capabilities and applications of these agents pose varying degrees of societal, ethical, and legal risks, necessitating a tailored approach that ensures proportional human oversight. Risk assessments in AI regulation are important in order to balance innovation and ethical responsibility, while formulating a global understanding of the societal risks posed by AI. But a key issue in how AI will shape our society remains; that is, the issue of centralization vs. decentralization (Brynjoflsson and NG, 2023). Centralized regulatory models, while offering streamlined governance and clear accountability, risk concentrating power and control in the hands of a few entities—whether they are governments, corporations, or other stakeholders. This centralization may lead to a lack of transparency, unequal access to regulatory oversight mechanisms, and the exclusion of marginalized voices from critical decision-making processes. Moreover, reliance on centralized authorities increases vulnerability to systemic risks, such as data monopolization, regulatory capture, or single points of failure in AI governance. On the other hand, decentralized frameworks present a compelling alternative for promoting equitable participation, trust, and resilience in AI regulation. Therefore, beyond adopting global standards for risk classification, we seek to leverage blockchain technology in an attempt promote decentralized governance and equitable access to AI through our ETHOS framework. 

\section{The ETHOS Model}
Embedding decentralized governance at the core of AI regulation can help us move toward a future that prioritizes inclusivity, mitigates risks of power concentration, and enables all citizens to shape how AI impacts their lives. This vision aligns with our broader goal of balancing innovation with ethical responsibility while ensuring that no single entity dominates the trajectory of AI development and deployment. The ETHOS framework thus positions blockchain as the foundation for a more equitable, participatory, and resilient AI governance ecosystem. 

\subsection{Technological Foundations}
To implement a scalable, decentralized governance model, our proposed framework leverages the following core technologies:

\begin{itemize}
    \item \textbf{Smart Contracts}: Smart contracts are self-executing agreements stored on the blockchain that automate compliance enforcement, risk monitoring, and decision-making processes. For instance, they trigger actions like adjusting risk tiers, enforcing penalties, or revoking compliance certifications based on predefined benchmarks. By minimizing human intervention, smart contracts ensure transparency, efficiency, and trust in regulatory execution (Buterin, 2014).

    \item \textbf{Decentralized Autonomous Organizations (DAOs)}: DAOs form the backbone of ETHOS governance, enabling participatory decision-making through consensus mechanisms. Stakeholders—such as developers, regulators, auditors, and ethicists—vote on governance actions, including updates to risk thresholds, ethical guidelines, or approvals for high-risk AI agents. Governance decisions are logged immutably on the blockchain for transparency and accountability (Hassan and De Filippi, 2021).
\item \textbf{Oracles}: Oracles bridge off-chain and on-chain data by securely gathering, validating, and transmitting real-world information—such as performance logs, user feedback, and societal impact metrics—onto the blockchain (Hamda Al-Breiki et al., 2020). This ensures that ETHOS remains dynamic and responsive to AI agent performance while preventing data manipulation through decentralized verification and redundancy mechanisms.

    \item \textbf{Self-Sovereign Identity (SSI)}: SSI enables privacy-preserving and verifiable identity management for AI agents (Chaffer and Goldston, 2022). Each agent is assigned a digital identity containing compliance credentials, performance logs, and audit results. SSI ensures that sensitive metadata remains encrypted and accessible only to authorized stakeholders while allowing seamless validation of compliance records.
    \item \textbf{Soulbound Tokens (SBTs)}: SBTs act as non-transferable compliance certifications issued when AI agents meet predefined ethical benchmarks, such as bias mitigation, privacy safeguards, or transparency audits (Weyl et al., 2022). Breaches of compliance trigger smart contracts to flag or revoke SBTs, ensuring continuous accountability and ethical alignment.
    \item \textbf{Zero-Knowledge Proofs (ZKPs)}: ZKPs are cryptographic techniques that allow compliance verification without revealing sensitive data (Lavin et al., 2024). For example, ZKPs enable auditors to confirm that an AI agent meets ethical standards (e.g., bias mitigation) without accessing underlying datasets or proprietary algorithms. This ensures data confidentiality while maintaining trust in compliance outcomes.
    \item \textbf{Dynamic Risk Classification System}: ETHOS uses a real-time risk classification system powered by blockchain and oracles to assess AI agent risk levels based on their autonomy, decision-making complexity, adaptability, and societal impact. Smart contracts continuously monitor agent performance against global and regional benchmarks, recalibrating risk profiles as new data emerges.
    \item \textbf{Immutable and Transparent Audit Trails}: Blockchain-based audit trails record every AI agent's decision, input, and outcome in immutable blocks. Each transaction includes: Input Data: Raw data (e.g., medical records, case files) used for decision-making. Decision Pathways: Algorithms, parameters, and logical reasoning employed. Output Results: The final decision, action, or recommendation. Verification Signatures: Cryptographic hashes ensuring data authenticity. This creates a tamper-proof, transparent mechanism for real-time monitoring and accountability.
    \item \textbf{Reputation-Based Systems}: Reputation systems assess the reliability and trustworthiness of validators, auditors, and AI agents within the ETHOS framework. Stakeholders earn reputation scores based on consistent, verified, and ethical participation, while malicious behavior (e.g., false verification or data manipulation) results in penalties or reduced reputation.
    \item \textbf{Tokenization and Staking Mechanisms}: ETHOS uses native tokens to incentivize validators and auditors for accurate compliance verification. Validators stake tokens to participate in risk assessments, creating a financial deterrent against false verification. Successful verification earns rewards, while malicious behavior results in token slashing.
\end{itemize}

\textbf{Problem:} Dangers associated with ineffective centralization of AI regulation and governance. 

\textbf{Solution:} A Global Registry framework, which leverages decentralized technologies to address concerns of ineffective centralization (Cihon et al., 2020). This framework establishes a global platform for AI agent registration, risk classification, and compliance monitoring. By utilizing blockchain technology, the ETHOS Global Registry advocates for immutable recordkeeping, automated compliance checks via smart contracts, and real-time updates to AI agent risk profiles based on validated off-chain data collected through oracles.

\subsection{ETHOS Governance}
To address concerns about centralized control, the ETHOS framework advocates for the use of DAOs to establish a transparent, participatory, and scalable governance structure. This decentralized framework empowers a diverse set of stakeholders, including governments, developers, ethicists, auditors, civil society groups, and end-users, to actively contribute to regulatory decision-making.

A core feature of DAOs is their reliance on weighted voting mechanisms (Fan et al., 2023). For instance, subject matter experts, such as ethicists in AI bias or medical professionals in healthcare AI, may carry greater weight in decisions relevant to their domain. At the same time, reputation scores, earned through consistent and trustworthy participation in the governance process, could further incentivize accountability and discourage malicious behavior. All governance actions—including the approval of high-risk AI agents, the adjustment of compliance thresholds, or the revocation of non-compliant systems—are permanently recorded on the blockchain, creating an immutable, transparent audit trail that fosters trust and accountability. 

Ultimately, DAOs show promise in fostering adaptive oversight by ensuring governance structures remain responsive to the dynamic and evolving nature of AI agents. Smart contracts automate decision enforcement, such as implementing updated compliance standards or triggering escalated oversight for flagged systems. It is important to evalaute whether this automation can reduce delays, minimize human intervention, and set up conditions for seamless execution of regulatory decisions in this context.

\subsection{Identity Management of AI Agents on ETHOS}
To mitigate risks of data centralization and safeguard privacy, the ETHOS framework advocates for the incorporation of selective transparency mechanisms powered by SSI and SBTs. Indeed, SSI provides a privacy-preserving solution for managing AI agent credentials and compliance records (Haque et al., 2024). For example, each AI agent is assigned a unique SSI, which securely contains its risk-tier classification and compliance credentials, such as risk assessments based on the ETHOS framework’s four-tier system (unacceptable, high, moderate, minimal), third-party audits, and ethical compliance approvals. Examples include certifications aligned with regional and global standards, such as GDPR for data privacy (Naik and Jenkins, 2020) and FHIR or HIPAA for healthcare data security (Broshka and Hamid Jahankhani, 2024). SSI ensures verifiable identity management while preventing unauthorized access to sensitive information, empowering stakeholders—such as regulators, developers, and auditors—to validate an AI agent’s compliance securely. Trusted Execution Environments (TEEs) can be leveraged to secure processing and generating metadata (cryptographic proofs, action logs) for all agent actions. As highlighted in recent work on Trusted Distributed Artificial Intelligence (TDAI), TEEs are foundational for enabling secure and reliable computations in distributed systems by ensuring end-to-end trust and protecting sensitive operations at scale (AKİF AĞCA et al., 2022). Thus, TEEs can be leveraged to secure processing and generating metadata (cryptographic proofs, action logs) for all agent actions. The TEE outputs metadata for AI actions, which is linked to the agent's SSI. The SSI acts as the anchor, tying the AI agent’s identity to its actions and compliance records. SSI acts as the anchor, tying the AI agent’s identity to its actions and compliance records. 

Each AI agent’s SSI includes the following key metadata and compliance details. A global AI agent ID, functioning as a digital passport, uniquely identifies the AI agent across jurisdictions and regulatory bodies, ensuring traceability and accountability throughout its lifecycle. Immutable, on-chain records of the AI agent's operational performance, including accuracy metrics (i.e., data on the agent’s task success rates and error margins), bias mitigation (i.e., evidence of the agent’s fairness across demographic or contextual variables), and societal impact (i.e., etrics assessing the agent’s broader effects on individuals, communities, and the environment). These logs are cryptographically secured and transparently verifiable, allowing auditors and regulators to validate performance benchmarks dynamically. Verified outcomes of independent regulatory audits and inspections, including third-party compliance certifications (e.g., GDPR, HIPAA, etc.) and ethical benchmarks (e.g., transparency audits, privacy safeguards, and bias mitigation standards). These results are logged immutably on the blockchain and linked to the SSI, ensuring transparency and trust in the auditing process.

Complementing SSI, SBTs act as non-transferable, on-chain compliance certifications tied to predefined ethical benchmarks, such as bias mitigation, privacy safeguards, and transparency audits (Weyl et al., 2022). Certifying bodies issue SBTs when AI agents successfully meet compliance milestones, enabling instant verification of adherence to regulatory and ethical guidelines. In the event of a breach—such as the detection of bias or privacy violations—smart contracts automatically trigger actions to flag or revoke SBTs, restricting further deployment until corrective measures are taken. This automated enforcement mechanism ensures continuous accountability while minimizing manual intervention. 

To further preserve privacy, zero-knowledge proofs (ZKPs) enable compliance verification without exposing sensitive underlying data. ZKPs allow auditors or regulators to confirm that an AI agent meets ethical and regulatory benchmarks without revealing proprietary information, such as technology stacks or operational algorithms. Additionally, metadata—such as details of a developer’s proprietary systems or operational performance—remains encrypted and accessible exclusively to authorized stakeholders, ensuring data confidentiality. 

By combining SSI, SBTs, smart contracts, and ZKPs, the ETHOS framework embeds ethical compliance mechanisms that are both transparent and privacy-preserving. Examples of compliance credentials—such as risk-tier classifications, third-party audits, and certifications like GDPR and HIPAA ensure a jurisdictionally recognized and consistent standard for verification. This approach can empower equitable collaboration across global stakeholders while safeguarding sensitive AI agent data. Selective transparency ensures that compliance verification remains robust, verifiable, and automated, while sensitive operational details are shielded from misuse. As a result, ETHOS creates a resilient, accountable, and privacy-conscious governance system that addresses the challenges of centralization and promotes the responsible integration of AI agents into society.

\subsection{ETHOS in Practice}
The registry integrates a dynamic risk classification system powered by decentralized oracles and smart contracts, ensuring real-time, transparent, and adaptive AI governance. Blockchain serves as the foundation for securely aggregating and sharing AI agent performance data, enabling continuous updates to risk profiles based on real-world inputs. Oracles play a critical role in bridging off-chain and on-chain data by securely gathering, validating, and transmitting diverse data streams—such as task performance, societal impact metrics, and user feedback—onto the blockchain. These data streams can originate from Internet of Things (IoT) sensors, Application Programming Interfaces (APIs), or manual inputs, ensuring broad, real-time monitoring capabilities. 

To maintain accuracy and reliability, multiple oracles participate in decentralized verification processes, cross-referencing data sources to ensure consistency before anchoring information to the blockchain. For instance, oracles can validate resource allocation reports generated by AI agents against real-world outcomes, such as healthcare delivery logs or environmental monitoring results. This redundancy and consensus mechanism minimizes the risk of data manipulation or inaccuracies. Once securely recorded, smart contracts automatically compare on-chain data to predefined global and regional compliance benchmarks, facilitating continuous risk assessment and compliance monitoring. This includes recalibrating risk profiles, automating tier adjustments, triggering penalties for ethical or performance deviations, or escalating issues to human oversight when necessary. By continuously feeding and verifying new data, the system ensures proportional oversight that evolves in response to an AI agent’s real-world performance and societal impact. The decentralized, automated nature of this framework enhances transparency, accountability, and resilience while mitigating risks associated with centralized control, such as bottlenecks, regulatory capture, or security vulnerabilities.

Incentive structures are central to enhancing the decentralization, reliability, and scalability of the ETHOS framework, ensuring the integrity of data verification and compliance processes. The system relies on decentralized participants—validators and auditors—who assess the accuracy and integrity of real-time data submitted by oracles. This multi-layered process fosters trust and equitable participation across diverse stakeholders, including data providers (e.g., IoT devices, APIs), validators, and decision-makers such as regulators and AI developers.

The incentive mechanism operates through a well-defined workflow: Oracles gather and submit real-world data—such as performance logs, societal impact metrics, and user feedback—to the blockchain, where validators perform decentralized verification (Pasdar et al., 2022). Validators assess data for accuracy using techniques such as timestamp checks, cross-referencing with alternative sources, or consensus mechanisms like Proof of Stake (PoS) and Proof of Authority (PoA) (Bahareh Lashkari and Musilek, 2021; Manolache et al., 2022). By requiring majority consensus to approve submissions (Alhejazi and Mohammad, 2022), the system could minimize the risk of inaccurate or manipulated data entering the registry. A native token system underpins the incentive structure. Validators stake tokens to participate in the verification process, creating a financial deterrent against dishonesty—false verification or inaccurate assessments result in penalties, such as token slashing or temporary bans. Successful verification earns validators tokenized rewards, incentivizing ongoing participation and maintaining system scalability. Additionally, a reputation-based system assigns scores to validators based on their performance, rewarding consistently accurate verification with higher rewards or increased voting power in governance decisions. This approach fosters long-term reliability and trust within the ecosystem. Smart contracts enforce these incentive mechanisms automatically, ensuring transparency and accountability. Non-compliance or unethical behavior—such as approving fraudulent data or breaching ethical standards—triggers pre-programmed penalties, including freezing AI agent deployments or revoking developer credentials. This automated enforcement process aims to minimize human intervention while maintaining fairness and integrity.

\section{Decentralized Justice}
In a world where AI agents will have unprecedented influence over critical decisions, ensuring fairness, accountability, and transparency in their governance demands a justice system as dynamic as the technologies it seeks to regulate. Towards this end, the ETHOS framework offers a decentralized approach to justice, designed to address the unique challenges of AI governance while upholding ethical and legal principles in an increasingly automated society. Decentralized justice is built upon digital courts that leverage blockchain technology as the cornerstone for ensuring transparency, accountability, and fairness in resolving disputes in the digital age. That is, through game theory and mechanism design (Ast and Deffains, 2021), settling disputes becomes a matter of designing judicial architectures that leverage economic incentives to crowdsource jurors, enabling peer-driven judicial decisions facilitated by smart contracts (Aouidef et al., 2021). These mechanisms play a vital role in addressing disputes arising from the increasing autonomy and complexity of AI agents, setting the stage for broader questions of legal liability.

In the coming age of AI agents, leveraging decentralized justice mechanisms will be critical in ensuring a fair and just resolution of disputes, where impartiality and transparency are paramount. While decentralized justice provides the foundational framework for transparency and fairness in governance, its implementation relies heavily on innovative mechanisms for resolving disputes effectively and equitably. By harnessing the power of blockchain technology and integrating innovative economic models, decentralized justice systems can adapt to the complexities of AI-driven societies, ensuring that governance frameworks remain resilient, inclusive, and ethically sound. 

‌\subsection{Decentralized Dispute Resolution} 
Imagine a future where an autonomous AI agent is tasked with coordinating a humanitarian relief effort. Operating independently to optimize resource distribution, the AI agent prioritizes large urban centers where logistics are simpler and cost-effective, while deprioritizing smaller, harder-to-reach communities. This decision can unintentionally marginalize vulnerable populations and disregards fundamental human values such as fairness and equity.
While no legal precedents yet exist for such scenarios involving AI agents, past cases highlight the importance of fairness and accountability in disaster response. For instance, in the Flint Water Crisis, lawsuits resulted in a 626 million dollar settlement for victims after predominantly African American communities were disproportionately harmed by contaminated water (Flint Water Cases, 2022). Similarly, in the ongoing case Strickland, et al. v. United States Department of Agriculture (2022), Texas farmers allege that disaster and pandemic relief funds were distributed based on race and sex, rather than need (Strickland v. USDA, 2024). These cases underscore the critical need for equitable resource distribution, a challenge that will only grow as autonomous systems become integral to autonomous crisis management.
Disputes related to regulatory or operational non-compliance in AI governance can arise from various sources, such as disagreements over risk classifications, alleged breaches of ethical or legal standards, or conflicts between stakeholders. A cornerstone of DAO-driven dispute resolution is the transparent filing of disputes on a decentralized platform. As the ETHOS framework proposes that all metadata related to AI agent decision-making—such as inputs, outputs, and operational parameters—be recorded immutably on the blockchain, this could provide a complete and tamper-proof record of the AI agent's actions leading up to the dispute. Then, stakeholders can access verifiable evidence of what occurred, eliminating ambiguities and disputes over the facts. This immutable record fosters trust among all parties, enabling fair and evidence-based resolutions. 

With a network of independent verifiers, ETHOS tackles challenges of integrity and accuracy of the data before it is used in the dispute resolution process. Verifiers, selected based on SBT-proven reputation, analyze the blockchain-stored metadata to validate claims about the AI agent’s decision-making processes. 

This decentralized approach, in theory, may prevent any single entity from exerting undue influence. Once the verifiers establish the accuracy of the data and the resolution process is completed, smart contracts automatically execute the decision. These contracts can enforce outcomes such as financial compensation, penalties, or modifications to the AI agent's operational parameters. For instance, if an AI agent is found to have operated in a manner inconsistent with ethical standards, a smart contract could adjust its access to data, impose fines, or mandate updates to its programming. This automation eliminates delays and the need for human intervention, ensuring resolutions are implemented consistently, efficiently, and impartially. While decentralized mechanisms streamline dispute resolution, addressing the broader implications of AI agent liability requires an equally innovative and adaptive approach to accountability.

‌\subsection{Evolution of Legal Liability}

A holistic model for AI risk management is not built solely on technical and operational safeguards but must also address how AI agents may pose fundamental risks to questions of justice—an essential pillar of our society. An emerging issue with AI agents is the question of legal liability. AI agents pose distinct liability challenges that necessitate a structured accountability framework to address issues such as information asymmetry, complex value chains, and delegation of discretion (Dunlop et al., 2024; Soder et al, 2024). To address these challenges, we must answer the question of how responsibility is assigned, provide clarity in liability, and propose mechanisms for compensating damages caused by autonomous systems. Lior (2019) argues that AI agents are not entities that are capable of assuming legal responsibility for any wrongdoing as they lack judgement and are merely used as an instrument by the human overseer (Lior, 2019). We acknowledge the merits of his position, and while AI agents may currently lack the human qualities—such as judgment, intent, and moral responsibility—needed to assume legal accountability, it is not difficult to imagine a scenario where AI agents obtain such qualities and, as a result, have complicating effects for traditional conceptions of accountability.

ETHOS anticipates scenarios where legal ambiguities arise and offers mechanisms to ensure adherence to core legal values, such as proportional accountability, ethical transparency, and equity, even in uncharted territories. Indeed, the ETHOS framework addresses the challenges of legal liability by proposing innovative mechanisms that adapt traditional accountability structures to the unique characteristics of AI agents. Central to this is the concept of AI-specific legal entities, granting highly autonomous AI systems a limited form of legal status, akin to corporate entities (Doomen, 2023). This conceptual status envisions scenarios where AI agents could assume liability for damages caused by their actions, effectively shifting the burden of responsibility from developers and operators to the AI system itself. However, this approach is not without its worrisome implications—it raises profound ethical, legal, and societal questions about the nature of responsibility, intent, and the moral agency of AI systems. As creators of AI, we must tread carefully in considering such paths, weighing the long-term impacts of these decisions on societal structures and values. Recognizing these complexities, the ETHOS framework proposes a roadmap for exploring alternative scenarios and carefully guiding discussions about the evolving role of AI agents within the legal system. Rather than endorsing the immediate application of these ideas, this approach encourages thoughtful deliberation, offering a conceptual lens to consider how legal systems might adapt to address novel issues arising from the integration of AI agents into our society.

One approach worthy of consideration is the idea of AI-specific legal entities being governed by a decentralized governance body and required to maintain mandatory insurance coverage, ensuring financial compensation for damages and incentivizing risk mitigation through reduced premiums for safer, more ethically aligned systems. For instance, high-risk AI agents, such as those used in healthcare, require stringent measures, including legal entity registration, mandatory insurance, and frequent audits to ensure compliance. Moderate-risk agents, operating in less sensitive environments, would still mandate insurance and operator accountability, with periodic audits to confirm adherence to ethical and safety standards. Minimal-risk agents, by contrast, would have lighter oversight, with general liability remaining with operators and optional insurance coverage. Accountability is ultimately at the hands of developers and operators first, who remain responsible for ensuring ethical design, avoiding biases, and adhering to safety standards. Certification processes and mandatory audits would help enforce compliance while legal mechanisms apportion shared liability in cases of joint fault. To handle disputes arising from AI agent operations, specialized dispute resolution mechanisms such as dedicated tribunals and alternative dispute resolution (ADR) processes can be employed. 

In addition to developer and operator accountability, we propose extending the concept of AI agent insurance to the consumer level, allowing individual users to insure AI agents operating within their personal or professional domains. This innovation would not only empower users to manage liability risks associated with their AI agents but also foster the emergence of a specialized insurance industry tailored to the unique challenges posed by AI technologies. For example, a consumer's AI agent personal assistant might mistakenly execute a high-value financial transaction outside of preset parameters, resulting in significant financial losses that could be mitigated through consumer-level AI agent insurance. Policies could be customized based on the agent's autonomy and application, opening avenues for dynamic risk assessment and mitigation strategies. Consumer-level insurance offers a dual advantage: it alleviates some of the liability burdens on developers and operators while providing an additional safeguard to encourage public trust and adoption of AI systems.

The ETHOS framework has the potential to facilitate the implementation of these accountability mechanisms by providing a structured and adaptive governance model tailored to the unique challenges of AI agents. By leveraging decentralized governance structures, such as DAOs (Decentralized Autonomous Organizations), ETHOS provides transparent oversight and participatory decision-making for AI-specific legal entities. Its dynamic risk classification system enables proportional oversight, aligning mandatory insurance requirements and liability standards with the potential societal impact of AI agents. Furthermore, the framework's use of blockchain-based transparency tools and audit trails ensures that compliance processes remain robust, verifiable, and accountable. By embedding these principles into its design, the ETHOS framework creates the necessary infrastructure to operationalize developer, operator, and consumer-level accountability, fostering trust and ethical alignment in AI systems.

\subsection{A Collaborative Approach to Regulating AI Agents}
Strategies for AI governance cannot be created in a vacuum. The transformative potential of AI agents will likely affect every facet of society, making it imperative to draw on diverse perspectives and foster collaboration. Ethical governance of AI agents will require a collaborative effort (Gianni et al., 2022), marked by societal engagement, public education, and innovative policy development to ensure inclusivity, adaptability, and global alignment (Celso Cancela-Outeda, 2024). Stakeholder engagement through, for example, public consultations, is critical to this framework's success (Kallina and Singh, 2024), incentivizing active participation from governments, industry leaders, ethicists, and the public to shape regulatory policies that reflect diverse perspectives and societal priorities (Celso Cancela-Outeda, 2024). It will be important to assess if our ETHOS framework has an effect on creating a sense of collective ownership over AI governance, possibly by fostering transparency and building trust. Transparency in policy development through open processes further enhances societal trust and buy-in, resulting in more robust and acceptable policies (Matasick, 2017). 

Education can empower individuals and communities to engage with AI agents (Kim et al., 2022). AI literacy programs should educate the public about the capabilities, limitations, and ethical considerations of AI, enabling informed decision-making and advocacy. Simplified explanations, ethical awareness, and hands-on interaction with AI agents can foster familiarity and reduce fear of the unknown. We advocate for educators to consider how AI is framed when communicating to students (Kim, 2023). That is, framing of AI agents as collaborators with humans, rather than adversaries or replacements, could, in theory, significantly influence regulatory approaches and societal perceptions. This paradigm, which we have previously described (Chaffer et al., 2024), emphasizes the symbiotic relationship between humans and AI, fostering shared accountability in the human-agent coevolution. We extend our paradigm to explore how collaborative framing—wherein multiple stakeholders share their assumptions, values, and goals—can reduce societal resistance to AI adoption and foster the development of human-centric systems designed to enhance user capabilities rather than operate independently. This approach can inspire policies that balance innovation with ethical safeguards, focusing on shared outcomes rather than restrictive measures. Another active approach is the establishment of regulatory sandboxes, which provides a practical solution for testing AI agents under controlled conditions, allowing developers, regulators, and other stakeholders to assess potential risks, benefits, and unintended consequences before real-world deployment (Díaz-Rodríguez et al., 2023). Insights from these environments can refine oversight measures, ensuring that regulation remains proportional and effective. By incorporating diverse perspectives through participatory governance and fostering trust through transparency and education, regulation can reflect collective principles of humanity while safeguarding fundamental rights and values.

Given the global nature of AI development and deployment (Ren and Du, 2024), international coordination is vital to promoting harmonized regulations across jurisdictions (Cihon et al., 2019). Collaboration between nations can help establish common standards and guidelines, reducing fragmentation and ensuring consistent governance of AI agents. Such alignment could facilitate cross-border innovation while mitigating risks like regulatory arbitrage (Lancieri et al., 2024.), where developers exploit less stringent rules in certain regions. Furthermore, continuous evaluation will be crucial to maintaining the relevance and effectiveness of the regulatory framework. As AI agents  evolve and new applications emerge, periodic reviews are necessary to reassess risk profiles, update regulatory requirements, and address unforeseen challenges (Barrett et al., 2023). This iterative process can help the regulatory strategy to adapt to technological advancements and societal shifts, enabling the safe and equitable integration of AI agents across domains.

Ultimately, the approach to regulating AI agents must balance the interests of governments, the public, and innovators to responsibly harness the transformative potential of AI. Frameworks such as risk-based regulation, accountability mechanisms, and collective governance principles can enable a flexible and adaptive approach, ensuring that innovation is not stifled but aligned with societal goals. In this way, AI regulation becomes not merely a constraint but a guiding force, channeling technological advancements to serve the greater good while ensuring fairness, transparency, and human dignity. Anticipatory studies such as this one, though speculative in nature, play a crucial role in addressing foreseeable risks associated with AI agents. By engaging with the ethical, societal, and legal challenges of autonomous systems before they fully materialize, these studies help inform proactive regulatory strategies, ensuring that AI agents are seamlessly integrated into society as partners in progress. We are hopeful that this collaborative and forward-looking strategy can play a part in ensuring that AI agents become partners in progress, and seamlessly integrated into society while upholding principles of ethical responsibility and societal trust.

\section{Discussion}
The ETHOS framework represents a conceptual exploration aimed at provoking thoughtful dialogue about AI regulation, recognizing that its real-world implementation requires empirical validation and broad stakeholder collaboration. This paper underscores the importance of empirical validation through pilot programs and interdisciplinary collaboration to refine the ETHOS framework into a practical tool for AI governance. To that end, our conceptual analysis of AI agents hinged on a fundamental question: what are we truly dealing with when we engage with rational agents capable of independent reasoning, learning, and decision-making? AI agents challenge foundational ideas about the autonomy, rationality, and ethics of intelligent machines. These entities are not merely tools; they are autonomous systems capable of reasoning, decision-making, and iterative learning. This challenges traditional notions of agency, as AI agents can be thought of as having beliefs, desires, and intentions. As a result, they demand new frameworks for understanding the boundaries of accountability, ethical responsibility, and societal integration.

Building on this philosophical understanding of the challenges AI agents pose, the ETHOS model represents a novel approach to AI regulation. Its uniqueness lies in its ability to integrate philosophical anticipation of AI agents' diverse use cases into a structured, dynamic risk-based framework, while also proposing an adaptive, decentralized regulatory approach that evolves alongside AI technology. The ETHOS framework acknowledges that AI agents will not be static entities. Their diverse capabilities and contexts of deployment require a model that is as dynamic as the technologies it seeks to regulate. Traditional regulatory approaches often fall short in this regard, relying on rigid, one-size-fits-all mechanisms that fail to accommodate the complexity and variability of AI applications. In contrast, the ETHOS model leverages blockchain technology as its core infrastructure to address these limitations.

The recent opinion by the European Data Protection Board (EDPB) on the use of personal data in the development and deployment of AI models underscores the critical need to balance innovation with the protection of fundamental rights under the General Data Protection Regulation (GDPR) (EDPB, 2024). The EDPB’s three-step test for evaluating legitimate interest, particularly its focus on necessity and proportionality in data processing, aligns closely with the principles embedded in the ETHOS framework. ETHOS operationalizes these principles through dynamic compliance tools such as blockchain-based audit trails and privacy-preserving technologies like Zero-Knowledge Proofs (ZKPs), ensuring that data processing adheres to the necessity test while respecting the data minimization principle. Moreover, ETHOS introduces a structured and scalable approach to the balancing test via its tiered risk classification system and decentralized governance model, enabling transparent assessment of how controllers' interests align with the rights and expectations of data subjects. By leveraging participatory mechanisms, including DAOs, ETHOS directly addresses the EDPB’s call for context-aware evaluations of data subjects' reasonable expectations. These innovations establish the ETHOS framework as a complementary tool to GDPR guidelines, offering a robust and forward-looking solution for the complexities of AI governance in a rapidly evolving technological landscape.

However, it is important to mention that our paper on the ETHOS framework faces several limitations that could impact its effectiveness in regulating AI agents. Its reliance on tiered risk categorization risks oversimplifying the nuanced nature of AI applications, as many agents may exhibit hybrid risk profiles and could be part of swarms (i.e, a group of AI agents). Implementation challenges arise from the difficulty of achieving global consensus on ethical principles and governance standards, which may lead to fragmented adoption. While this paper is aspirational and takes a systems-thinking approach, a lot of the conceptual work attempts to bridge many elements together that may be incompatible or require extensive optimization in real-world conditions. Indeed, the absence of empirical validation leaves questions about the framework's real-world applicability, but it could be tested through pilot programs in specific industries—such as finance—where outcomes can be measured for ethical compliance, risk mitigation, and societal impact. Such testing would provide valuable insights for policymakers refining regulations, developers optimizing AI system designs, and end users benefiting from safer, more transparent AI technologies.

Our conceptual work is uniquely positioned as a keystone paper in this emerging field of study. As this is an emerging field, it is difficult to account for all the works being published, particularly papers with ambitious aims such as ours, which are typically posted on open repositories (i.e., arXiv, SSRN, etc.) or published as conference papers. Very recent and important papers include work by Chan et al. (2024), which explores key mechanisms to facilitate the governance of increasingly autonomous AI agents, emphasizing the importance of visibility in ensuring accountability and oversight. The study identifies three primary visibility mechanisms: agent identifiers, real-time monitoring, and activity logs (Chan et al., 2024a). Adler et al. (2024) explore the concept of personhood credentials as a means to enhance trust and security on online platforms, particularly in the face of increasingly sophisticated AI systems (Adler et al., 2024). Furthermore, Chan et al. (2024b) propose a framework for assigning IDs to AI system instances to enhance accountability and accessibility, particularly in impactful settings like financial transactions or human interactions, while acknowledging adoption incentives, implementation strategies, and associated risks. Domkundwar et al. (2024) propose and evaluate three safety frameworks—an LLM-powered input-output filter, a safety agent, and a hierarchical delegation system with embedded checks—to mitigate risks in AI systems collaborating with humans, demonstrating their effectiveness in enhancing safety, security, and responsible deployment of AI agents (Domkundwar et al., 2024). On matters related to societal integration, Bernardi and colleagues (2024) highlight the importance of defensive AI systems, international cooperation, and targeted investments to enhance societal resilience, secure adaptation, and mitigate risks posed by advanced AI technologies (Bernardi et al., 2024). Furthermore, to assist with societal integration and risk management, Gipiškis et al. (2024) present a comprehensive catalog of risk sources and management measures for general-purpose AI (GPAI) systems, spanning technical, operational, and societal risks across development, training, and deployment stages (Gipiškis et al. (2024), offering a neutral and publicly accessible resource to assist stakeholders in AI governance and standards development. 

Our paper contributes to a new field called "Decentralized AI Governance", which explores the integration of decentralized technologies—such as blockchain, smart contracts, and DAOs—to regulate, monitor, and ensure the ethical alignment of increasingly autonomous AI systems (Kaal, 2024). Recently, Krishnamoorthy (2024) introduced a novel governance framework for Artificial General Intelligence (AGI) that integrates Human-in-the-Loop (HITL) approaches with blockchain technology to ensure robust and adaptable oversight mechanisms, thereby helping establish a new academic field of AI governance through blockchain. It is critical to mention that Gabriel Montes and Ben Goertzel (2019) are widely regarded as pioneers in decentralized AI (Montes and Goertzel, 2019), and that their vision of democratizing access to AI systems and ensuring ethical alignment through decentralized infrastructure has been instrumental in shaping this emerging field. 

\section{Conclusion}

Our main contribution to the literature lies in the conceptualization and introduction of the ETHOS (Ethical Technology and Holistic Oversight System) framework, a global, decentralized governance model specifically tailored to regulate AI agents. Unlike existing works that focus on isolated mechanisms or risk management strategies, our work provides a comprehensive and unified approach to address the unique challenges posed by AI agents—autonomous systems capable of reasoning, learning, and adapting. The ETHOS framework leverages Web3 technologies such as blockchain, smart contracts, DAOs, SSI, and SBTs to establish a decentralized global registry for AI agents. By integrating risk-based categorization (unacceptable, high, moderate, minimal) with proportional oversight mechanisms, ETHOS bridges the gap between philosophical understandings of AI agents' autonomy, decision-making complexity, adaptability, and impact potential, and practical strategies for governance. This approach ensures dynamic risk classification, real-time monitoring, automated compliance enforcement, and enhanced transparency, while addressing concerns around centralization, accountability, and societal integration. Therefore, by combining philosophical, technical, and operational insights into a cohesive model, ETHOS aims to lay the foundation for a resilient, inclusive, and adaptive governance system, contingent upon successful implementation, empirical validation, and broad adoption across diverse stakeholders, which could in turn meet the demands of an AI-driven future.
\section*{Acknowledgments}
We would like to thank the Valhalla group for their helpful discussions in the preparation of this manuscript. We acknowledge the use of Chat Generative Pre-Trained Transformer (ChatGPT), developed by OpenAI, in assisting with the drafting, refinement, and editing throughout the paper. The researchers did not receive any funding for this study. The authors hold the belief that the rule of law is the cornerstone of a just society. This work seeks to extend these enduring principles to address the novel challenges posed by AI agents, ensuring that innovation is aligned with societal values.

\section*{References}

Adler, S., Hitzig, Z., 1, Jain, S., Brewer, C., Chang, W., Diresta, R., Lazzarin, E., Mcgregor, S., Seltzer, W., Siddarth, D., Soliman, N., South, T., Sporny, M., Srivastava, V., Bailey, J., Christian, B., Critch, A., Duffy, K., and Ho, E. (2024). Personhood credentials: Artificial intelligence and the value of privacy-preserving tools to distinguish who is real online. https://openreview.net/pdf?id=pEYxSx0frs
Ağca, M. A., Faye, S., and Khadraoui, D. (2022). A survey on trusted distributed artificial intelligence. IEEE Access, 10, 55308-55337.

Albahri, A. S., Duhaim, A. M., Fadhel, M. A., Alhamzah Alnoor, Baqer, N. S., Laith Alzubaidi, O. S., ... Muhammet Deveci. (2023). A systematic review of trustworthy and explainable artificial intelligence in healthcare: Assessment of quality, bias risk, and data fusion. Information Fusion, 96, 156–191. https://doi.org/10.1016/j.inffus.2023.03.008

Allen, G.C., and Adamson, G. (2024). The AI Safety Institute International Network. Next Steps and Recommendations. https://csis-website-prod.s3.amazonaws.com/s3fs-public/2024-10/241030AllenSafetyNetwork.pdf?VersionId=v3oOqfbw384pBCjL05.qVf28ArGoBUYT

Aouidef, Y., Ast, F., and Deffains, B. (2021). Decentralized Justice: A Comparative Analysis of Blockchain Online Dispute Resolution Projects. Frontiers in Blockchain, 4. https://doi.org/10.3389/fbloc.2021.564551

.Ast, F., and Deffains, B. (2021, June 30). When Online Dispute Resolution Meets Blockchain: The Birth of Decentralized Justice. Stanford Journal of Blockchain Law and Policy. https://stanford-jblp.pubpub.org/pub/birth-of-decentralized-justice/release/1

Balayogi, G., Lakshmi, A. V., and Sophie, S. L. (2025). Human-Centric Ethical AI in the Digital World. In Ethical Dimensions of AI Development (pp. 175-196). IGI Global.

Barrett, A. M., Newman, J., Nonnecke, B., Hendrycks, D., Murphy, E. R., and Jackson, K. (2023). AI Risk-Management Standards Profile for General-Purpose AI Systems (GPAIS) and Foundation Models. Center for Long-Term Cybersecurity. https://cltc.berkeley.edu/wp-content/uploads/2023/11/Berkeley-GPAIS-Foundation-Model-Risk-Management-Standards-Profile-v1.0.pdf

Bipartisan House Task Force On Artifical Intelligence (2024). Final Report. https://www.speaker.gov/wp-content/uploads/2024/12/AI-Task-Force-Report-FINAL.pdf 

Beer, J. M., Fisk, A. D., and Rogers, W. A. (2014). Toward a framework for levels of robot autonomy in human-robot interaction. Journal of Human-Robot Interaction, 3(2), 74. https://doi.org/10.5898/jhri.3.2.beer

Bengio, Y., Hinton, G., Yao, A., Song, D., Abbeel, P., Darrell, T., ... Dragan, A. (2024). Managing extreme AI risks amid rapid progress. Science, 384(6698), 842–845. https://doi.org/10.1126/science.adn0117

Bernardi, J., Mukobi, G., Greaves, H., Heim, L., and Anderljung, M. (2024). Societal Adaptation to Advanced AI. ArXiv.org. https://arxiv.org/abs/2405.1029

Broshka, E., and Hamid Jahankhani. (2024). Evaluating the Importance of SSI-Blockchain Digital Identity Framework for Cross-Border Healthcare Patient Record Management. Advanced Sciences and Technologies for Security Applications, 87–110. https://doi.org/10.1007/978-3-031-72821-15

Brundage, M., Avin, S., Clark, J., Toner, H., Eckersley, P., Garfinkel, B., ... Lyle, C. (2018). The malicious use of artificial intelligence: Forecasting, prevention, and mitigation. ArXiv.org. https://doi.org/10.17863/cam.22520

Brynjolfsson, E., and Ng, A. (2023). Big AI can centralize decision-making and power, and that’s a problem. Missing links in AI governance, 65.

Buterin, V. (2014). Ethereum White Paper: A next-generation smart contract and decentralized application platform. Ethereum Foundation.

Card, D., and Smith, N. A. (2020). On Consequentialism and Fairness. Frontiers in Artificial Intelligence, 3. https://doi.org/10.3389/frai.2020.00034‌

Cihon, P., Maas, M. M., and Kemp, L. (2020). Should Artificial Intelligence Governance be Centralised? Proceedings of the AAAI/ACM Conference on AI, Ethics, and Society. https://doi.org/10.1145/3375627.3375857

Celso Cancela-Outeda. (2024). The EU’s AI Act: A framework for collaborative governance. Internet of Things, 27, 101291. https://doi.org/10.1016/j.iot.2024.101291

Cervantes, J.-A., López, S., Rodríguez, L.-F., Cervantes, S., Cervantes, F., and Ramos, F. (2019). Artificial Moral Agents: A Survey of the Current Status. Science and Engineering Ethics, 26(2), 501–532. https://doi.org/10.1007/s11948-019-00151-x

Chaffer, T. J., and Goldston, J. (2022). On the Existential Basis of Self-Sovereign Identity and Soulbound Tokens: An Examination of the “Self” in the Age of Web3. Journal of Strategic Innovation and Sustainability, 17(3). https://articlearchives.co/index.php/JSIS/article/view/5855

Chaffer, T. J., Goldston, J., and D. A. T. A. I, G.  (2024). Incentivized Symbiosis: A Paradigm for Human-Agent Coevolution. ArXiv.org. https://arxiv.org/abs/2412.06855

Chan, A., Ezell, C., Kaufmann, M., Wei, K., Hammond, L., Bradley, H., Bluemke, E., Rajkumar, N., Krueger, D., Kolt, N., Heim, L., and Anderljung, M. (2024a). Visibility into AI Agents. ArXiv.org. https://doi.org/10.1145/3630106.3658948

Chan, A., Ai, A., Kolt, N., Wills, P., Anwar, U., Schroeder De Witt, C., Rajkumar, N., Hammond, L., Krueger, D., Heim, L., and Anderljung, M. (2024b.). IDs for AI Systems. ArXiv.org. https://arxiv.org/pdf/2406.12137

Council of Europe (2024). Council of Europe Framework Convention on Artificial Intelligence and Human Rights, Democracy and the Rule of Law. https://rm.coe.int/1680afae3c

Dennis, L., Fisher, M., Slavkovik, M., and Webster, M. (2016). Formal verification of ethical choices in autonomous systems. Robotics and Autonomous Systems, 77, 1–14.

Díaz-Rodríguez, N., Ser, J. D., Coeckelbergh, M., López, M., Herrera-Viedma, E., and Herrera, F. (2023). Connecting the dots in trustworthy artificial intelligence: From AI principles, ethics, and key requirements to responsible AI systems and regulation. Information Fusion, 99, 101896. https://doi.org/10.1016/j.inffus.2023.101896

Doomen, J. (2023) The artificial intelligence entity as a legal person. Information and Communciations Technology Law, 23(3) 277-287 

Domkundwar, I., Superagi, M., and Superagi, I. (2024). Safeguarding AI Agents: Developing and Analyzing Safety Architectures. https://arxiv.org/pdf/2409.03793

Dunlop, C., Pan, W., Smakman, J., Soder, L., Swaroop, S., and Kolt, N. Position: AI Agents and Liability–Mapping Insights from ML and HCI Research to Policy. Workshop on Socially Responsible Language Modelling Research.

European Data Protection Board (EDPB) (2024). Opinion 28/2024 on certain data protection aspects related to the processing of personal data in the context of AI models. https://www.edpb.europa.eu/system/files/2024-12/edpbopinion202428ai-modelsen.pdf

Elster, J. (1982). Rationality. La philosophie contemporaine/Contemporary philosophy: Chroniques nouvelles/A new survey, 111-131.

Fan, C., Zhang, C., Yahja, A., and Mostafavi, A. (2019). Disaster City Digital Twin: A vision for integrating artificial and human intelligence for disaster management. International Journal of Information Management, 56, 102049–102049. https://doi.org/10.1016/j.ijinfomgt.2019.102049

Fan, Y., Zhang, L., Wang, R., and Imran, M. A. (2023). Insight into voting in DAOs: conceptual analysis and a proposal for evaluation framework. IEEE Network, 38(3), 92-99.

Ferber, D., Omar, Wölflein, G., Wiest, I. C., Clusmann, J., Leßman, M.-E., Foersch, S., Lammert, J., Tschochohei, M., Jäger, D., Salto-Tellez, M., Schultz, N., Truhn, D., and Kather, J. N. (2024). Autonomous Artificial Intelligence Agents for Clinical Decision Making in Oncology. ArXiv.org. https://arxiv.org/abs/2404.04667

Flint Water Cases, No. 16-cv-10444 (E.D. Mich. 2022). https://www.mied.uscourts.gov/index.cfm?pageFunction=CasesOfInterest 

Gianni, R., Lehtinen, S., and Nieminen, M. (2022). Governance of responsible AI: From ethical guidelines to cooperative policies. Frontiers in Computer Science, 4. https://doi.org/10.3389/fcomp.2022.873437

Gipiškis, R., San, A., Shen, Z., 1, C., Regenfuß, A., Gil, A., and Holtman, K. (2024). Risk Sources and Risk Management Measures in Support of Standards for General-Purpose AI Systems. ArXiv.org. https://arxiv.org/pdf/2410.23472

Global Partnership on Artificial Intelligence (2024). https://gpai.ai/

Greenblatt, R., Denison, C., Wright, B., Roger, F., MacDiarmid, M., Marks, S., Treutlein, J., Belonax, T., Chen, J., Duvenaud, D., Khan, A., Michael, J., Mindermann, S., Perez, E., Petrini, L., Uesato, J., Kaplan, J., Shlegeris, B., Bowman, S. R., and Hubinger, E. (2024). Alignment faking in large language models. ArXiv.org. https://arxiv.org/abs/2412.14093

Guo, T., Chen, X., Wang, Y., Chang, R., Pei, S., Chawla, N. V., Wiest, O., and Zhang, X. (2024). Large Language Model based Multi-Agents: A Survey of Progress and Challenges. ArXiv.org. https://arxiv.org/abs/2402.01680

Hamda Al-Breiki, Muhammad, Salah, K., and Davor Svetinovic. (2020). Trustworthy Blockchain Oracles: Review, Comparison, and Open Research Challenges. IEEE Access, 8, 85675–85685. https://doi.org/10.1109/access.2020.2992698

Hassan, S., and De Filippi, P. (2021). Decentralized Autonomous Organization. Internet Policy Review, 10(2). https://doi.org/10.14763/2021.2.1556

Haque, R. U., Hasan, A. S. M. T., Ali, M., Zhang, Y., and Xu, D. (2024). SSIFL, Self sovereign identity based privacy preserving federated learning. Journal of Parallel and Distributed Computing, 191, 104907 https://doi.org/10.1016/j.jpdc.2024.104907

Hess, A., and Kjeldsen, J. E. (Eds.). (2024). Ethos, Technology, and AI in Contemporary Society: The Character in the Machine. Taylor and Francis.

Hinton, G. (2024) Noble Minds. https://www.nobelprize.org/nobel-minds/

Hooker, J., and Kim, T. W. (2019). Truly autonomous machines are ethical. AI Magazine, 40(4), 66–73. https://doi.org/10.1609/aimag.v40i4.2863

Jiang, B., Xie, Y., Wang, X., Su, W. J., Taylor, C. J., and Mallick, T. (2024). Multi-modal and multi-agent systems meet rationality: A survey. ICML 2024 Workshop on LLMs and Cognition. https://arxiv.org/html/2406.00252v2S7

Kaal, W. A. (2024). AI Governance Via Web3 Reputation System. SSRN Electronic Journal. https://doi.org/10.2139/ssrn.4941807

Kallina, E., and Singh, J. (2024). Stakeholder involvement for responsible AI development: A process framework. Proceedings of the ACM Conference on Fairness, Accountability, and Transparency, 1–14. https://doi.org/10.1145/3689904.3694698

Kim, J. (2023). Leading teachers’ perspectives on teacher-AI collaboration in education. Education and Information Technologies, 29(7), 8693–8724. https://doi.org/10.1007/s10639-023-12109-5

Kim, J., Lee, H., and Cho, Y. H. (2022). Learning design to support student-AI collaboration: Perspectives of leading teachers for AI in education. Education and Information Technologies, 27(5), 6069–6104. https://doi.org/10.1007/s10639-021-10831-6

Krishnamoorthy, M. V. (2024). Enhancing Responsible AGI Development: Integrating Human-in-the-loop Approaches with Blockchain-based Smart Contracts. Journal of Advances in Mathematics and Computer Science, 39(9), 14–39. https://doi.org/10.9734/jamcs/2024/v39i91924

Lashkari, B. and Musilek, P. (2021). A Comprehensive Review of Blockchain Consensus Mechanisms. IEEE Access, 9, 43620–43652. https://doi.org/10.1109/access.2021.3065880

Lavin, R., Liu, X., Mohanty, H., Norman, L., Zaarour, G., and Krishnamachari, B. (2024). A Survey on the Applications of Zero-Knowledge Proofs. ArXiv.org. https://arxiv.org/abs/2408.00243

Laitinen, A., and Sahlgren, O. (2021). AI Systems and Respect for Human Autonomy. Frontiers in Artificial Intelligence, 4. https://doi.org/10.3389/frai.2021.705164

Lancieri, F., Edelson, L., and Bechtold, S. (2024). AI Regulation: Competition, arbitrage, and regulatory capture. ETH Library. https://doi.org/10.3929/ethz-b-000708626

Lior, A. (2019). AI entities as AI agents: Artificial intelligence liability and the AI respondeat superior analogy. Mitchell Hamline L. Rev., 46, 1043.

Lungu, M. A. (2024). Smart Urban Mobility: The Role of AI in Alleviating Traffic Congestion. Proceedings of the International Conference on Business Excellence, 18(1), 1441–1452. https://doi.org/10.2478/picbe-2024-0118

Malek Mechergui, and Sarath Sreedharan. (2024). Goal Alignment: Re-analyzing Value Alignment Problems Using Human-Aware AI. Proceedings of the AAAI Conference on Artificial Intelligence, 38(9), 10110–10118. https://doi.org/10.1609/aaai.v38i9.28875

Malik, F. H., and Lehtonen, M. (2015). A review: Agents in smart grids. Electric Power Systems Research, 131, 71–79. https://doi.org/10.1016/j.epsr.2015.10.004

Manolache, M. A., Sergiu Manolache, and Nicolae Tapus. (2022). Decision Making using the Blockchain Proof of Authority Consensus. Procedia Computer Science, 199, 580–588. https://doi.org/10.1016/j.procs.2022.01.071

Matasick, C. (2017). Open government: How transparency and inclusiveness can reshape public trust. OECD Publishing. https://www.oecd-ilibrary.org/docserver/9789264268920-8-en.pdf

McGraw, D. K. (2024). Ethical Responsibility in the Design of Artificial Intelligence (AI) Systems. International Journal on Responsibility, 7(1). https://doi.org/10.62365/2576-0955.1114
Alhejazi, M. M., and Mohammad, R. M. A. (2022). Enhancing the blockchain voting process in IoT using a novel blockchain Weighted Majority Consensus Algorithm (WMCA). Information Security Journal: A Global Perspective, 31(2), 125-143.

Montes, G. A., and Goertzel, B. (2019). Distributed, decentralized, and democratized artificial intelligence. Technological Forecasting and Social Change, 141, 354–358. https://doi.org/10.1016/j.techfore.2018.11.010

Naik, N., and Jenkins, P. (2020). Your Identity is Yours: Take Back Control of Your Identity Using GDPR Compatible Self-Sovereign Identity. Aston Publications Explorer (Aston University), 1–6. https://doi.org/10.1109/besc51023.2020.9348298

National Institute of Standards and Technology (2024). Artificial intelligence risk management framework: Generative artificial intelligence profile. National Institute of Standards and Technology. https://doi.org/10.6028/nist.ai.600-1

Pasdar, A., Lee, Y. C., and Dong, Z. (2022). Connect API with Blockchain: A Survey on Blockchain Oracle Implementation. ACM Computing Surveys, 55(10), 1–39. https://doi.org/10.1145/3567582

Pedron, S. M., and da Cruz, J. D. A. (2020). The future of wars: Artificial intelligence (AI) and lethal autonomous weapon systems (LAWS). International Journal of Security Studies, 2(1), 2.

Raso, F. A., Hilligoss, H., Krishnamurthy, V., Bavitz, C., and Kim, L. (2018). Artificial intelligence and human rights: Opportunities and risks. SSRN Electronic Journal. https://doi.org/10.2139/ssrn.3259344

Regulation - EU - 2024/1689 - EN - EUR-Lex. (2024). Europa.eu. https://eur-lex.europa.eu/legal-content/EN/TXT/?uri=CELEX:32024R1689

Reiling, D. (2020). Courts and artificial intelligence. International Journal for Court Administration, 11(2). https://doi.org/10.36745/ijca.343

Ren, Q., and Dui, J. (2024). Harmonizing innovation and regulation: The EU Artificial Intelligence Act in the international trade context. Computer Law and Security Review, 54, 106028. https://doi.org/10.1016/j.clsr.2024.106028

Shin, W., Bu, S.-J., and Cho, S.-B. (2019). Automatic financial trading agent for low-risk portfolio management using deep reinforcement learning. ArXiv.org. https://arxiv.org/abs/1909.03278

Soder, L., Smakman, J., Dunlop, C., Pan, W., and Swaroop, S. Levels of Autonomy: Liability in the age of AI Agents. In Workshop on Socially Responsible Language Modelling Research.
Strickland v. United States Dep’t of Agric., 2:24-CV-60-Z https://clearinghouse.net/case/45369/ 

Swanepoel, D., and Corks, D. (2024). Artificial intelligence and agency: Tie-breaking in AI decision-making. Science and Engineering Ethics, 30(2). https://doi.org/10.1007/s11948-024-00476-2
United Nations, (2024). Governing AI for Humanity: Final Report https://www.un.org/sites/un2.un.org/files/governingaiforhumanityfinalreporten.pdf

Uriel, D., and Remolina, N. (2024). Artificial intelligence at the bench: Legal and ethical challenges of informing—or misinforming—judicial decision-making through generative AI. Data and Policy, 6. https://doi.org/10.1017/dap.2024.53

Wasil, A. R., Clymer, J., Krueger, D., Dardaman, E., Campos, S., and Murphy, E. R. (2024). Affirmative safety: An approach to risk management for high-risk AI. ArXiv.org. https://arxiv.org/abs/2406.15371

Weyl, E. G., Puja Ohlhaver, and Vitalik Buterin. (2022). Decentralized Society: Finding Web3’s Soul. SSRN Electronic Journal. https://doi.org/10.2139/ssrn.4105763

Winfield, A. F. T., and Jirotka, M. (2018). Ethical governance is essential to building trust in robotics and artificial intelligence systems. Philosophical Transactions of the Royal Society A: Mathematical, Physical and Engineering Sciences, 376(2133), 20180085. https://doi.org/10.1098/rsta.2018.0085

Xi, Z., Chen, W., Guo, X., He, W., Ding, Y., Hong, B., Zhang, M., Wang, J., Jin, S., Zhou, E., Zheng, R., Fan, X., Wang, X., Xiong, L., Zhou, Y., Wang, W., Jiang, C., Zou, Y., Liu, X., and Yin, Z. (2023). The Rise and Potential of Large Language Model Based Agents: A Survey. ArXiv.org. https://arxiv.org/abs/2309.07864

Xia, Y., Shin, S.-Y., and Lee, H.-A. (2024). Adaptive Learning in AI Agents for the Metaverse: The ALMAA Framework. Applied Sciences, 14(23), 11410. https://doi.org/10.3390/app142311410

Xu, L., Hu, Z., Zhou, D., Ren, H., Dong, Z., Keutzer, K., Ng, S.-K., and Feng, J. (2024). MAgIC: Investigation of Large Language Model Powered Multi-Agent in Cognition, Adaptability, Rationality and Collaboration. Proceedings of the 2021 Conference on Empirical Methods in Natural Language Processing, 7315–7332. https://doi.org/10.18653/v1/2024.emnlp-main.416

Zhang, Z. J., Schoop, E., Nichols, J., Mahajan, A., and Swearngin, A. (2024). From Interaction to Impact: Towards Safer AI Agents Through Understanding and Evaluating UI Operation Impacts. ArXiv.org. https://arxiv.org/abs/2410.09006

\end{document}